\documentclass[sigconf]{acmart}
\usepackage{multirow}
\AtBeginDocument{%
  \providecommand\BibTeX{{%
    \normalfont B\kern-0.5em{\scshape i\kern-0.25em b}\kern-0.8em\TeX}}}

\copyrightyear{2022} 
\acmYear{2022} 
\setcopyright{acmcopyright}
\acmConference[MM '22]{Proceedings of the 30th ACM International Conference on Multimedia}{October 10--14, 2022}{Lisboa, Portugal}
\acmBooktitle{Proceedings of the 30th ACM International Conference on Multimedia (MM '22), October 10--14, 2022, Lisboa, Portugal}
\acmPrice{15.00}
\acmDOI{10.1145/3503161.3548541}
\acmISBN{978-1-4503-9203-7/22/10}

\begin{document}

\title{MMRotate: A Rotated Object Detection Benchmark \\using PyTorch}

\author{Yue Zhou*}
\orcid{0000-0002-3080-6721}
\affiliation{%
  \institution{Department of EE, Shanghai Jiao Tong University}
  \city{Shanghai}
  \country{China}}
\email{sjtu_zy@sjtu.edu.cn}
\thanks{* Equal contribution. \dag Correspondence author. Gefan Zhang is also with Cowa Robot, Co Ltd, Wuhu, Anhui, China}

\author{Xue Yang*}
\orcid{0000-0002-7084-9101}
\affiliation{%
  \institution{MoE Key Lab of Artificial Intelligence, Shanghai Jiao Tong University}
  \city{Shanghai}
  \country{China}}
\email{yangxue-2019-sjtu@sjtu.edu.cn}

\author{Gefan Zhang}
\orcid{0000-0003-3519-3976}
\affiliation{%
  \institution{MoE Key Lab of Artificial Intelligence, Shanghai Jiao Tong University}
  \city{Shanghai}
  \country{China}
}
\email{lizaozhouke@sjtu.edu.cn}

\author{Jiabao Wang}
\orcid{0000-0003-4951-3088}
\affiliation{%
  \institution{Northwestern Polytechnical University}
  \city{Xi'an}
  \country{China}
}
\email{jbwang@mail.nwpu.edu.cn}

\author{Yanyi Liu}
\orcid{0000-0002-6123-9328}
\affiliation{%
  \institution{Northeastern University}
  \city{Shenyang}
  \country{China}
}
\email{2001688@stu.neu.edu.cn}

\author{Liping Hou}
\orcid{0000-0002-4984-3290}
\affiliation{%
  \institution{University of Chinese Academy of Sciences}
  \city{Beijing}
  \country{China}
}
\email{houliping17@mails.ucas.ac.cn}

\author{Xue Jiang \dag}
\orcid{0000-0001-7099-6817}
\affiliation{%
  \institution{Department of EE, Shanghai Jiao Tong University}
  \city{Shanghai}
  \country{China}
}
\email{xuejiang@sjtu.edu.cn}

\author{Xingzhao Liu}
\orcid{0000-0002-4533-3904}
\affiliation{%
  \institution{Department of EE, Shanghai Jiao Tong University}
  \city{Shanghai}
  \country{China}
}
\email{xzhliu@sjtu.edu.cn}

\author{Junchi Yan \dag}
\orcid{0000-0001-9639-7679}
\affiliation{%
  \institution{MoE Key Lab of Artificial Intelligence, Shanghai Jiao Tong University}
  \city{Shanghai}
  \country{China}
}
\email{yanjunchi@sjtu.edu.cn}

\author{Chengqi Lyu}
\orcid{0000-0002-6356-3922}
\affiliation{%
  \institution{Shanghai AI Laboratory}
  \city{Shanghai}
  \country{China}
}
\email{lvchengqi@pjlab.org.cn}

\author{Wenwei Zhang}
\orcid{0000-0002-2748-4514}
\affiliation{%
  \institution{Nanyang Technological University}
  \city{Singapore}
  \country{Singapore}
}
\email{wenwei001@ntu.edu.sg}

\author{Kai Chen}
\orcid{0000-0002-6820-2325}
\affiliation{%
  \institution{Shanghai AI Laboratory}
  \institution{SenseTime Research}
  \city{Shanghai}
  \country{China}
}
\email{chenkai@sensetime.com}

\renewcommand{\shortauthors}{Y. Zhou and X. Yang, et al.}

\begin{abstract}
We present an open-source toolbox, named MMRotate, which provides a coherent algorithm framework of training, inferring, and evaluation for the popular rotated object detection algorithm based on deep learning. MMRotate implements 18 state-of-the-art algorithms and supports the three most frequently used angle definition methods. To facilitate future research and industrial applications of rotated object detection-related problems, we also provide a large number of trained models and detailed benchmarks to give insights into the performance of rotated object detection. MMRotate is publicly released at https://github.com/open-mmlab/mmrotate.
\end{abstract}






\maketitle

\section{Introduction}
In recent years, deep learning has achieved tremendous success
in fundamental computer vision applications such as image recognition \cite{he2016deep}, object detection \cite{girshick2015fast,liu2016ssd,redmon2016you,ren2015faster,tian2019fcos} and image segmentation \cite{he2017mask,long2018textsnake}. In light of this, deep learning has also been applied to areas such as faces detection \cite{shi2018real}, text detection \cite{zhou2017east,jiang2017r2cnn,liu2018fots,ma2018arbitrary,liao2018textboxes++} and aerial images detection \cite{yang2018automatic,yang2019scrdet,yang2021r3det}. In these object detection tasks, oriented bounding boxes (OBBs) are widely used instead of horizontal bounding boxes (HBBs) because they can better align the objects for more accurate identification. This kind of special object detection is called rotated object detection, also known as arbitrary-oriented object detection. In addition to the three applications mentioned above, rotated object detection is also widely used in 3D objects detection \cite{zheng2020rotation} and retail scenes detection \cite{chen2020piou,pan2020dynamic}.

Different approaches utilize different angle definition methods, optimization strategies (e.g., optimizers, learning rate schedules, epoch numbers, and data augmentation pipelines), and CUDA operators (e.g. IoU and NMS for OBBs). To encompass the diversity of components used in various models, we have proposed the MMRotate toolbox covering recent popular rotated object detection approaches in a unified framework. The toolbox now implements 18 rotated object detection methods, 10 CUDA speed-up operators, and 12 losses. Integrating various algorithms confers code reusability and therefore dramatically simplifies the implementation of algorithms. Moreover, the unified framework allows different approaches to be compared against each other fairly and that their key effective components can be easily investigated. To the best of our knowledge, MMRotate supports most angle definition methods in various open source toolboxes, and it will facilitate future research on rotated object detection. 

MMRotate is hosted on GitHub under the Apache-2.0 License. The repository contains the compressed archive file of software\footnote{https://github.com/open-mmlab/mmrotate/archive/refs/heads/main.zip} and documentation, including installation instructions, dataset preparation scripts, API documentation\footnote{https://mmrotate.readthedocs.io/en/latest/}, model zoo, tutorials and user manual. MMRotate re-implements 18 state-of-the-art rotated object detection algorithms and provides extensive benchmarks and models trained on popular academic datasets. In addition to (distributed) training and testing scripts, It offers a rich set of utility tools covering visualization and demonstration. 

% 这周弄两个支持 onnx 的模型出来
% and deployment. The models provided by MMRotate are easily converted to onnx \cite{} which is widely supported by deployment frameworks and hardware devices. Therefore, it is useful for both academic researchers and industrial developers.

\section{RELATED WORK}
% Rotated object detection is an extended research direction of general object detection, which aims to predict more accurate bounding boxes and preserve the direction information of the object.
%介绍文字旋转检测发展历史, 水平框文字检测过渡到旋转框文字检测
\textbf{Text detection.} 
Text detection aims to localize the bounding boxes of text instances. Recent research focus has shifted to challenging arbitrary-shaped text detection \cite{he2017mask, jiang2017r2cnn, ma2018arbitrary, long2018textsnake, liao2020real}. R$^2$CNN \cite{jiang2017r2cnn} simultaneously predicts the axis-aligned and inclined boxes by adding an inclined box branch and uses an inclined NMS to obtain the detection results. While Mask R-CNN \cite{he2017mask} can be used to detect texts, it might fail to detect curved and dense texts due to the rectangle-based ROI proposals. On the other hand, RRPN \cite{ma2018arbitrary} proposes a Rotation RPN to generate inclined proposals with text orientation angle information and project arbitrary-oriented proposals to the feature map with Rotation ROI pooling. TextSnake \cite{long2018textsnake} describes text instances with a series of ordered, overlapping disks.

%介绍遥感检测发展历史，dota数据集，遇到的一些问题，目前的解决方案
\textbf{Aerial image detection.} 
Aerial image detection plays a vital role in the military and attracts more and more attention in civilian field \cite{ding2018learning,yang2020arbitrary,yang2021rethinking,yang2022on}. It aims to predict more accurate bounding boxes and preserve the direction information of the object on aerial images (including ship, plane, vehicles, bridge, etc.). Although rotated object detection provides more accurate prediction results than horizontal detection, it requires defining a new bounding box representation. The most common is the $\theta$-based representation $(x,y,w,h,\theta)$, and it adds an extra angle parameter based on the horizontal box. Depending on the angle range, it can be divided into OpenCV definition ($D_{oc}, \theta \in [-\pi/2,0)$) \cite{yang2018automatic,yang2019scrdet, yang2022scrdet++}, long edge 90° definition ($D_{le90}, \theta \in [-\pi/2, \pi/2)$) \cite{ding2018learning,han2021redet}, and long edge 135° definition ($D_{le135},\theta \in [-\pi/4, 3\pi/4)$) \cite{han2021align}. Recent works used two-dimensional Gaussian distribution \cite{yang2021rethinking,yang2021learning} and point sets \cite{xu2020gliding,pan2020dynamic,qian2021learning,ming2021dynamic} to represent objects, which have achieved excellent results. Feature alignment is another research direction of lifting rotated object detection performance. R$^3$Det \cite{yang2021r3det} proposes a feature refinement module to re-construct the feature map based on the refined bounding box output from the previous stage. S$^2$A-Net \cite{han2021align} proposes an alignment convolution to alleviate the misalignment between axis-aligned convolutional features and arbitrary oriented objects. Recently, ReDet \cite{han2021redet} began to study a novel rotation-equivariant RoI Align to produce rotation-equivariant features. Label Assignment is also a research hotspot. DAL \cite{ming2021dynamic} reconsiders whether IoU is a truly credible division basis and defines a new matching degree. SLA  \cite{ming2021sparse} proposes a sparse label assignment strategy to achieve training sample selection based on posterior IoU distribution. SASM \cite{hou2022shape} proposes two novel shape-adaptive strategies which can dynamically select samples and measure the quality of positive samples.

\begin{table}[!tb]
\centering
\caption{Open source rotated object detection benchmarks.}
\label{tab:benchmarks}
\resizebox{0.48\textwidth}{!}{
\begin{tabular}{c|ccccc}
\hline
Benchmark & AerialDet & JDet & OBBDet & AlphaRotate & MMRotate \\ \hline
DL library & PyTorch & Jittor & PyTorch & TensorFlow & PyTorch \\
\hline
Inference & \multirow{2}{*}{PyTorch} & \multirow{2}{*}{Jittor} & \multirow{2}{*}{PyTorch} & \multirow{2}{*}{TensorFlow} & PyTorch \\
 engine &  &  &  &  & \multicolumn{1}{l}{onnx runtime} \\
\hline
\multirow{2}{*}{OS} & \multirow{2}{*}{Linux} & Windows & Windows & \multirow{2}{*}{Linux} & Windows \\
 &  & Linux & Linux &  & Linux \\
\hline
Algorithm & 5 & 8 & 9 & 16 & 18 \\
Dataset & 1 & 4 & 5 & 11 & 4 \\
Doc. & - & - & - & \checkmark & \checkmark \\
Easy install & - & - & - & - & \checkmark \\
Maintain & - & \checkmark & \checkmark & \checkmark & \checkmark \\ \hline
\end{tabular}}
\end{table}

\begin{table*}[!tb]
\centering
\caption{Accuracy comparison of rotated object detectors on DOTA v1.0. MS means multiple scale image split. RR means random rotation. All models are inferred with one 2080Ti GPU.} 
\label{tab:dota}
\resizebox{0.95\textwidth}{!}{
\begin{tabular}{c|cccccccc}
\hline
Baseline                       & \textbf{Technique} & \multicolumn{1}{l}{fp16} & Box Def. & Lr schd. & Mem.(GB) & Inf. time (fps) & Aug.                  & mAP                       \\ \hline
\multirow{6}{*}{RetinaNet-H \cite{lin2017focal}}   & -                           & -                        & oc       & 1x       & 3.38     & 15.7            & -                     & 64.55                     \\
                               & GWD  \cite{yang2021rethinking}                       & -                        & oc       & 1x       & 3.39     & 15.5            & -                     & 69.55                     \\
                               & KFIoU \cite{yang2022kfiou}                      & -                        & le90     & 1x       & 3.38     & 15.1            & -                     & 69.60                     \\
                               & KFIoU \cite{yang2022kfiou}                      & -                        & oc       & 1x       & 3.39     & 15.6            & -                     & 69.76                     \\
                               & KFIoU \cite{yang2022kfiou}                       & -                        & le135    & 1x       & 3.38     & 15.3            & -                     & 69.77                     \\
                               & KLD  \cite{yang2021learning}                       & -                        & oc       & 1x       & 3.39     & 15.6            & -                     & 69.94                     \\ \hline
\multirow{8}{*}{RetinaNet-O \cite{lin2017focal}}   & -                           & -                        & le90     & 1x       & 3.38     & 16.9            & -                     & 68.42                     \\
                               & -                           & \checkmark                        & le90     & 1x       & 2.36     & 22.4            & - & \multicolumn{1}{l}{68.79} \\
                               & CSL  \cite{yang2020arbitrary}                         & \checkmark                          & le90     & 1x       & 2.60     & 24.0            & - & \multicolumn{1}{l}{69.51} \\
                               & KLD \cite{yang2021learning} & - & le90 & 1x & 3.35 & 16.9 & - & 70.22\\
                               & -                           & -                        & le135    & 1x       & 3.38     & 17.2            & -                     & 69.79                     \\
                               & ATSS \cite{zhang2020bridging}                           & -                        & le90    & 1x       & 3.12     & 18.2            & -                     & 70.64                     \\
                               & ATSS \cite{zhang2020bridging}                           & -                        & le135    & 1x       & 3.19     & 18.8            & -                     & 72.29                     \\
                               & -                           & -                        & le90     & 1x       & 3.78     & 17.5            & MS+RR                 & 76.50                     \\ \hline
\multirow{5}{*}{RepPoints \cite{yang2019reppoints}}     & -                           & -                        & oc       & 1x       & 3.45     & 15.6            & -                     & 59.44                     \\
                               & SASM \cite{hou2022shape}                       & -                        & oc       & 1x       & 3.53     & 15.7            & -                     & 66.45                     \\
                               & G-Rep \cite{hou2022grep}                      & -                        & le135    & 1x       & 4.05     & 8.6             & -                     & 69.49                     \\
                               & CFA \cite{pan2020dynamic}                         & -                        & le135    & 1x       & 3.45     & 16.1            & -                     & 69.63                     \\
                               & CFA \cite{pan2020dynamic}                        & -                        & oc       & 40e      & 3.45     & 16.1            & -                     & 73.45                     \\ \hline
\multirow{3}{*}{FCOS \cite{tian2019fcos}}     & -                           & -                        & le90       & 1x       & 4.18     &  20.9           & -                     & 70.70                     \\
& CSL \cite{yang2020arbitrary} & - & le90 & 1x & 4.23 & 20.2 & - & 71.76\\
& KLD \cite{yang2021learning} & - & le90 & 1x & 4.18 & 20.7 & - & 71.89\\
\hline
\multirow{4}{*}{R$^3$Det \cite{yang2021r3det}}      & -                           & -                        & oc       & 1x       & 3.54     & 12.4            & -                     & 69.80                     \\
                               & ATSS \cite{zhang2020bridging}                        & -                        & oc       & 1x         & 3.65 & 13.6            & -                     & 70.54   \\
                               & KLD \cite{yang2021learning}                         & -                        & oc       & 1x       & 3.54     & 12.4            & -                     & 71.83                     \\
                               & KFIoU  \cite{yang2022kfiou}                      & -                        & oc       & 1x       & 3.62     & 12.2            & -                     & 72.68                     \\ \hline
% \multirow{2}{*}{R$^3$Det tiny} &                             & -                        & oc       & 1x       & 3.23     & 15.6            & -                     & 70.18                     \\
                            %   & KLD                         & -                        & oc       & 1x       & 3.44     & 14.0            & -                     & 72.76                     \\ \hline
\multirow{2}{*}{S$^2$ANet \cite{han2021align}}     & -                           & -                        & le135    & 1x       & 3.14     & 15.5            & -                     & 73.91                     \\
                               & -                           & \checkmark                        & le135    & 1x       & 2.17     & 17.4            & - & \multicolumn{1}{l}{74.19} \\ \hline
\multirow{12}{*}{Faster RCNN \cite{ren2015faster}}  & -                           & -                        & le90     & 1x       & 8.46     & 16.5            & -                     & 73.40                     \\
                               & Gliding Vertex \cite{xu2020gliding}             & -                        & le90     & 1x       & 8.45     & 16.4            & -                     & 73.23                     \\
                              & Oriented RCNN \cite{xie2021oriented}              & \checkmark                        & le90     & 1x       & 7.37     & 21.2            & - & \multicolumn{1}{l}{75.63} \\
                               & Oriented RCNN  \cite{xie2021oriented}             & -                        & le90     & 1x       & 8.46     & 16.2            & -                     & 75.69                     \\
                               & RoI Trans. \cite{ding2018learning}                  & \checkmark                        & le90     & 1x       & 7.56     & 19.3            & -                     & 75.75                     \\
                               & ReDet  \cite{han2021redet}                     & \checkmark                        & le90     & 1x       & 7.71     & 13.3            & -                     & 75.99                     \\
                               & RoI Trans. \cite{ding2018learning}                  & -                        & le90     & 1x       & 8.67     & 14.4            & -                     & 76.08                     \\
                               & ReDet \cite{han2021redet}                      & -                        & le90     & 1x       & 9.32     & 10.9            & -                     & 76.68                     \\
                               & RoI Trans. \cite{ding2018learning} + Swin-T \cite{liu2021swin}       & -                        & le90     & 1x       & 9.23     & 10.9            & -                     & 77.51                     \\
                               & RoI Trans. \cite{ding2018learning}                  & -                        & le90     & 1x       & 8.96      & 14.4            & MS+RR                 & 79.66                     \\
                               & ReDet \cite{han2021redet}                      & -                        & le90     & 1x       & 9.63     & 10.9            & MS+RR                 & 79.87                     \\
                              & RoI Trans. \cite{ding2018learning} + KLD \cite{yang2021learning} + Swin-T \cite{liu2021swin} & - & le90 & 1x & 12.3 & 10.9 & MS+RR & 80.90\\
                               \hline
\end{tabular}}
\end{table*}

%介绍开源工具箱的历史
\textbf{Open source toolbox.} 
Several open source rotated object detection toolboxes have been developed over the years to meet the increasing demand from both academia and industry. 

AerialDetection\footnote{https://github.com/dingjiansw101/AerialDetection} is the pioneer of deep learning-based open source rotated object detection toolbox. It was publicly released in 2019, and provides evaluation tools for DOTA data set \cite{xia2018dota}. It supports five rotated object detection methods, e.g., Rotated RetinaNet \cite{lin2017focal}, Rotated Faster R-CNN \cite{ren2015faster} and RoI Trans \cite{ding2018learning}. However, it has not been maintained anymore. OBBDetection\footnote{https://github.com/jbwang1997/OBBDetection} \cite{xie2021oriented} is another popular open source oriented object detection toolbox. It supports 9 rotated object detection methods and provides a series of efficient processing tools for huge remote sensing images. AerialDetection and OBBDetection are both modified based on MMDetection\footnote{https://github.com/open-mmlab/mmdetection} \cite{chen2019mmdetection}, which is a state-of-the-art open source object detection toolbox based on PyTorch. Nevertheless, they cannot enjoy the latest technology provided by MMDetection, since they rely on a specific old version of MMDetection. JDet\footnote{https://github.com/Jittor/JDet} is an open source aerial image object detection toolbox based on a high-performance deep learning library \cite{hu2020jittor}. It can be deployed on multiple platforms such as Linux and Windows, and the 8 detectors it reproduces have faster inference speed than PyTorch. TensorFlow-based rotated object detection toolbox AlphaRotate\footnote{https://github.com/yangxue0827/RotationDetection} \cite{yang2021alpharotate} has been released recently. It currently implements 18 rotated object detection methods, including the algorithms that PyTorch and Jittor do not support, e.g., GWD \cite{yang2021rethinking}, KLD \cite{yang2021learning}, and R$^3$Det \cite{yang2021r3det}. Comprehensive comparisons among these open source toolboxes are given in Table \ref{tab:benchmarks}.

\section{ROTATED OBJECT DETECTION STUDIES}

%不同组件的对比实验

% backbone resnet 50, swin-tiny, PVT-v2
Many important factors can affect the performance of deep learning-based detectors. This section investigates the angle definition method, backbones, and loss of network architectures. We exchange the above-mentioned components between different rotated object detection approaches to measure the performance, memory usage, and inference time.

\textbf{Angle definition method.} OpenCV definition, long edge 90° definition, and long edge 135° definition are all supported by MMRotate. All rotated object detection algorithms can easily switch between these three angle definition methods by modifying the configuration file. Meanwhile, we established an angle conversion API to facilitate other angle definition methods.

\textbf{Backbone.} ResNet50 \cite{he2016deep} are commonly used in object detection approaches. To improve the accuracy, we also introduce a transformer-based backbone Swin transformer \cite{liu2021swin}. Table \ref{tab:dota} compares ResNet50 and Swin-T in terms of memory, inference time and mAP by plugging them in RoI Trans. It has been shown that Swin-T significantly outperforms ResNet50, although its inference speed is 21\% slower than that of ResNet50.

\textbf{Loss.} GWD, KLD, and KFIoU propose different loss to train the rotated object detector. Our experimental results in Table \ref{tab:dota} show that the KLD loss achieved the best mAP in the OpenCV definition method when using RetinaNet as the baseline. However, when using the R$^3$Det as the baseline, the KFIoU loss achieved the best mAP in the OpenCV definition method.

\textbf{Mixed precision training and Useful tools.} 
All detectors in MMRotate support mixed precision training. Our experimental results in Table \ref{tab:dota} show that the model trained with fp16 has a similar mAP as the original model. MMRotate also provides a range of efficient and convenient tools (including visualization, confusion matrix analysis, huge image inference), allowing researchers to focus on the rotated object detection algorithm itself.

% \begin{lstlisting}[language=python]
% import alpharotate
% from alpharotate.libs.models.detectors import retinanet
% from retinanet.build_whole_network import DetectionNetworkRetinaNet
% from configs import cfgs
% # init detector (e.g. RetinaNet)
% model = DetectionNetworkRetinaNet(cfgs=cfgs, is_training=True) 
% # training
% _, loss = model.build_whole_detection_network(input_img_batch=img, gt=gt) 
% \end{lstlisting}

\section{CONCLUSIONS}
With the practical importance and academic emergence for visual rotation detection, MMRotate is a deep learning benchmark for visual object rotation detection in PyTorch under the Apache-2.0 license. The architecture is designated for flexibility and ease of use to facilitate the deployment of rotated object detection in diverse domains, both in industrial applications and academic research. We will continue to improve the entire optimized benchmark and support representative detection methods in the future. We also welcome the community to participate in the development.

\begin{acks}
This work was supported by National Key Research and Development Program of China (2020AAA0107600), and National Natural Science Foundation of China (61971279, 61972250, 62022054).
\end{acks}

\bibliographystyle{ACM-Reference-Format}
\bibliography{sample-base}

\end{document}